\documentclass{article}

\usepackage{arxiv}

\usepackage[utf8]{inputenc} 
\usepackage[T1]{fontenc}    
\usepackage{hyperref}       
\usepackage{url}            
\usepackage{booktabs}       
\usepackage{amsfonts}       
\usepackage{nicefrac}       
\usepackage{microtype}      
\usepackage{lipsum}
\usepackage{graphicx}

\usepackage{booktabs}
\usepackage[load-configurations=version-1]{siunitx} 
 \usepackage{multirow}
 \usepackage{array}
 \usepackage{placeins}
 \usepackage{chngcntr}
 \usepackage[titletoc,title]{appendix}

\makeatletter

\title{Fairness in TabNet Model by Disentangled Representation for the Prediction of Hospital No-Show}

\author{%
Sabri Boughorbel \\
Qatar Computing Research Institute, Doha, Qatar \\
\texttt{sabri.boughorbel@gmail.com}
\And
Fethi Jarray 
University of Gabes, Medenine, Tunisia \\
\texttt{fjarray@gmail.com}\\
\And
Abdou Kadri \\
Sidra Medicine, Doha, Qatar \\
\texttt{akadri@sidra.org}\\
}

\begin{document}

\maketitle

\begin{abstract}
Patient no-shows is a major burden for health centers leading to loss of revenue, increased waiting time and deteriorated health outcome. Developing machine learning (ML) models for the prediction of no-shows could help addressing this important issue. It is crucial to consider fair ML models for no-show prediction in order to ensure equality of opportunity in accessing healthcare services. In this work, we are interested in developing deep learning models for no-show prediction based on tabular data while ensuring fairness properties. Our baseline model, TabNet, uses on attentive feature transformers and has shown promising results for tabular data. We propose Fair-TabNet  based on representation learning that disentangles predictive from sensitive components.  The model is trained to jointly minimize loss functions on no-shows and sensitive variables while ensuring that the sensitive and prediction representations are orthogonal. In the experimental analysis, we used a hospital dataset of 210,000 appointments collected in 2019. Our preliminary results show that the proposed Fair-TabNet improves the predictive, fairness performance and convergence speed over TabNet for the task of appointment no-show prediction. The comparison with the state-of-the art models for tabular data shows promising results and could be further improved by a better tuning of hyper-parameters.
\end{abstract}
\section{Introduction}
Patients who do not attend scheduled clinic appointments are referred to as no-shows. The latter is a common problem in clinics with a major impact on the economics, operation and health outcome of health centers. For example average rate of patient no-show  the U.S. is estimated to be 18.8 \% with a cost of 150 Billion in revenue loss \cite{gier2017missed, kheirkhah2015prevalence}. No-shows affect the recruitment of additional medical staff, disrupt appointment scheduling and extend patient waiting. These missed appointments are also causing health problems due to the interrupted or delayed follow-ups on patient conditions. Several approaches have been considered to tackle patient no-shows such as sending reminder, imposing sanctions or overbooking. Personalized reminders and intelligent overbooking are promising directions that are cost effective. These approaches require developing machine learning models that can perform individualized no-show prediction based on operational and health data. Several work have been published on the prediction of no-show using machine learning models \cite{carreras2020patient, li2019individualized, almuhaideb2019prediction}. The most commonly data features for no-show prediction can be categorized into: patient demographic, medical history, appointment details and patient behaviour \cite{carreras2020patient}. Such data is typically available in tabular format. Therefore traditional machine learning models such as decision trees are most commonly applied. It is attractive to develop deep learning models for tabular data.
\section{Related work}
We limit this section to references to previous work on fairness by representation learning via disentanglement.  \cite{louizos2015variational} proposed the Variational Fair Autoencoder (FVAE) where  the Independence between the target and sensitive variables was controlled by the Maximum Mean Discrepancy (MMD). The invariance was also controlled through generative adversarial network where one component maximizes the accuracy of the model and the other component minimizes the dependency between sensitive and target variables \cite{madras2018learning,zhang2018mitigating}. Creager et al. \cite{creager2019flexibly} proposed a fair disentangled  representation where  a set of sensitive variables and target variable is given at training but the true sensitive variable is only known at test time. Recently \cite{sarhan2020fairness} proposed an orthogonal disentangled representation that enforces the meaningful representation to be independent of the sensitive information. \cite{locatello2019fairness} showed that fairness can be improved through disentanglement representation.
\section{The Proposed Model: Fair-TabNet}
\begin{figure*}
\centering
\includegraphics[width=0.8\textwidth]{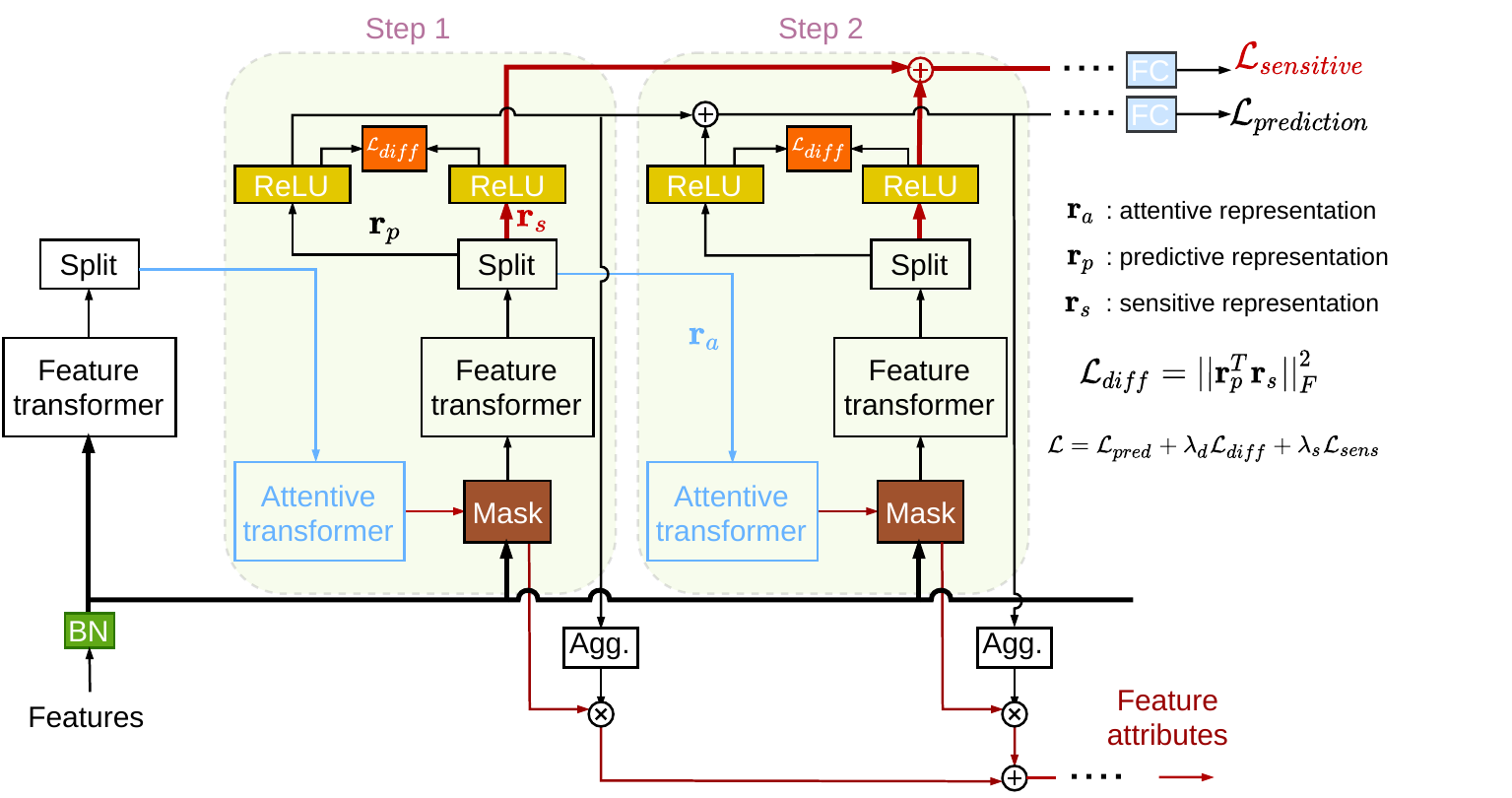} 
\caption{Architecture of the proposed Fair-TabNet Model. The red lines, $\mathcal L_{diff}$ and $\mathcal L_{sens}$ are the proposed extension. TabNet architecture is depicted in Figure \ref{fig:tabnet}.}
\label{fig:fair-tabnet}
\end{figure*}

Our goal is to develop a fair deep learning model for no-show classification suited for tabular data. Our baseline model is TabNet which has shown good predictive performance compared with the sate-of-the art models for tabular data \cite{arik2019tabnet}. The main building blocks in TabNet are: 1) Feature transformers, 2) Attentive transformers and 3) Masking.  We propose Fair-TabNet by introducing an additional component (${\bf r}_s$) in TabNet representation. ${\bf r}_s$ learns to correctly classify the sensitive variables on the training set. As depicted in Figure \ref{fig:fair-tabnet}, the loss $\mathcal L_{sens}$  ensures that ${\bf r}_s$ is a good representation of sensitive variables. A second loss term, $\mathcal L_{diff}=||\mathbf r_p^T \mathbf r_s||^2_F$, where $||\cdot||^2_F$ is the squared Frobenius norm, encourages orthogonality between sensitive ${\bf r}_s$ and the predictive representation ${\bf r}_p$. Hence this helps the disentanglement of both representations. The loss term $\mathcal L_{diff}$ is inspired by Deep Separation Networks \cite{bousmalis2016domain}.
 
Fair-TabNet is trained to minimize the weighted loss function $\mathcal{L}=\mathcal{L}_{pred}+\lambda_d\mathcal{L}_{sens}+\lambda_s\mathcal{L}_{diff}$ where $\mathcal{L}_{pred}$  is the binary cross-entropy loss on prediction targets (no-show), $\mathcal{L}_{sens}$  is a categorical cross entropy loss on the sensitive variables (gender and nationality). $\lambda_d$ and $\lambda_s$ are hyper-parameters that control the contribution of the introduced loss functions to the overall loss. 

\begin{table}
{\small
\begin{tabular}{lcc}
\hline
Metric      & AU-ROC         & AU-PRC         \\ \hline
LightGBM    & 77.88$\pm$0.19 & 46.3$\pm$0.46  \\ 
CatBoost    & 78.26$\pm$0.46 & 47.84$\pm$0.82 \\
XGBoost     & 78.74$\pm$0.42 & 47.8$\pm$0.77  \\
TabNet      & 75.93$\pm$1.78 & 43.83$\pm$3.83 \\
Fair-TabNet & 76.38$\pm$1.42 & 44.2$\pm$3.12  \\ \hline
\end{tabular}
}
\caption{Summary of Prediction performance. }
\label{tab:prediction_performance}
\end{table}


\section{Experiments}

\begin{figure}[t]
\centering
\includegraphics[width=0.48\textwidth]{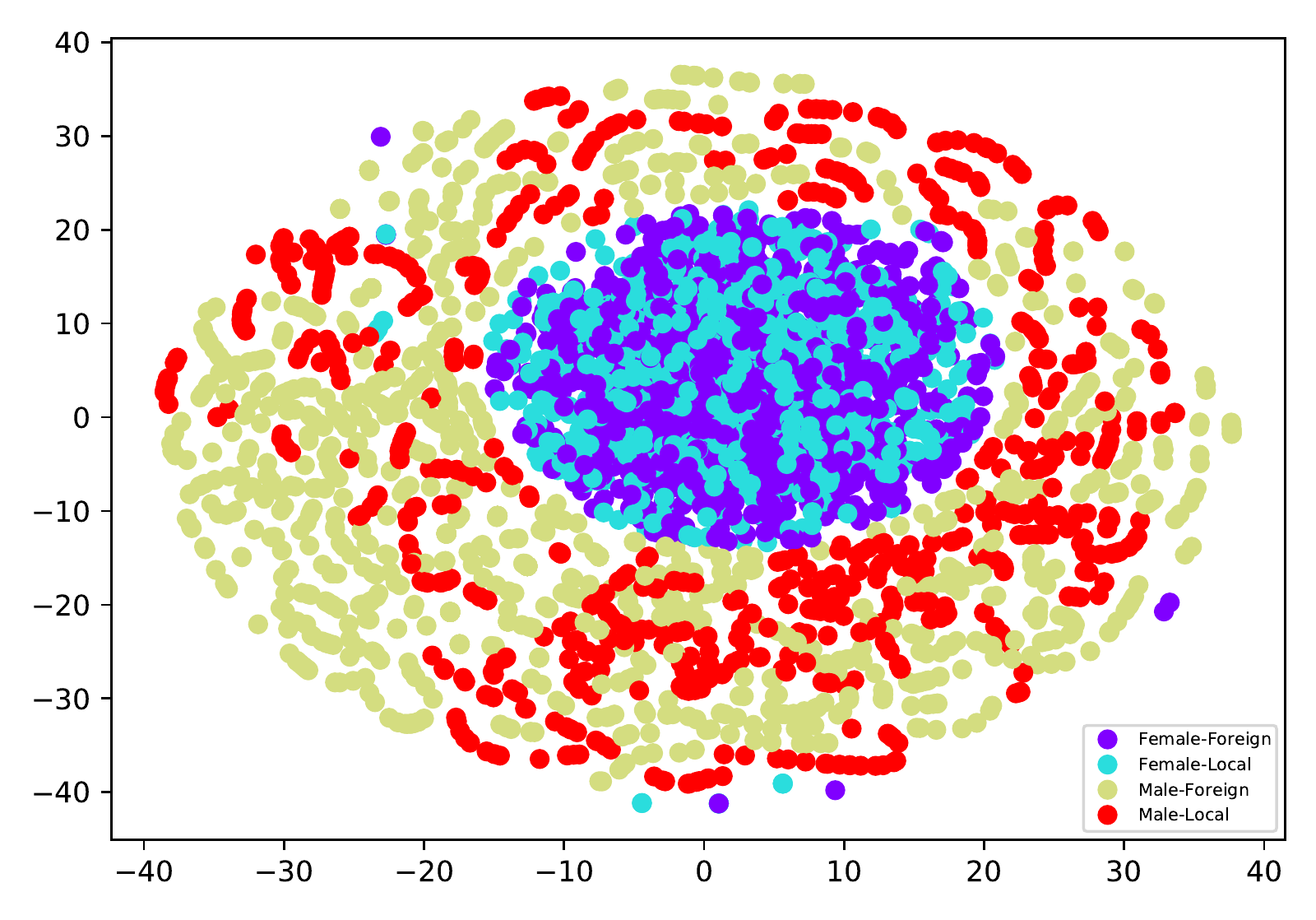}
    \caption{T-SNE plot of the sensitive representation ${\bf r}_s$}\label{fig:sens_repr}
\end{figure}

\begin{figure}[t]
\centering
\includegraphics[width=0.48\textwidth]{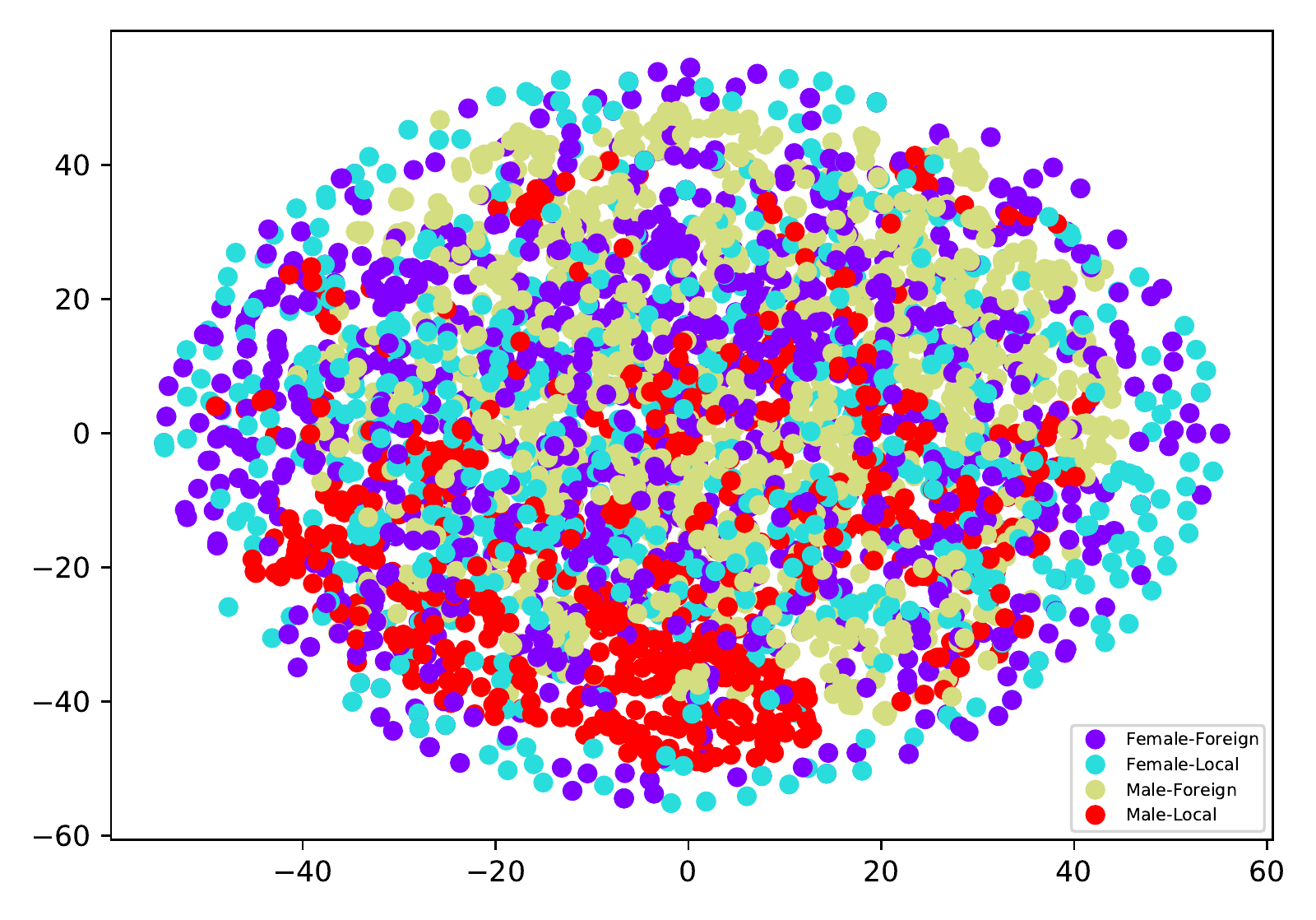}
\caption{T-SNE plot of the predictive representation ${\bf r}_p$}\label{fig:pred_repr}
\end{figure}

We used an appointment scheduling dataset from our hospital. The dataset, collected in 2019, gathered  211,028 appointments. The features are 1) patient information such as age, gender and nationality, 2) appointment information such as date, time, appointment duration, clinics, physician specialities, time from booking to appointment, new or follow-up appointment. In total the dataset included 42 features, in a tabular format, with numerical and categorical columns. The considered sensitive variables are Gender and Nationality. We are interested in group fairness. For Nationality variable we binarized the samples into local vs. foreigner. For the baseline TabNet, there are several hyper-parameters to choose for the experiments. We chose $n_p$=16 for the size of  $r_p$, $n_a=16$, for the size of $r_a$, $n_{steps}=5$ for the number of steps, $\gamma=1.5$. A better tuning of the hyper-parameters could lead to improvement in the predictive performance. For the proposed Fair-TabNet we chose the same parameters as in TabNet. The additional parameter in Fair-TabNet is $n_s=16$ for the size of the sensitive representation. The dataset is randomly split into training (70\%), validation (15\%) and test (15 \%) such that appointments of each day will be all in one of the three subsets. The splitting and models training is repeated 30 times  to estimate the mean and standard deviation of performance metrics. The chosen baseline methods for classification based on tabular are XGBoost, LightGBM and CatBoost.  Prediction performance is estimated using Area Under ROC (AU-ROC) and Area Under Precision-Recall curve (AU-PRC). The results are summarized in Table \ref{tab:prediction_performance}. Typically incorporating fairness properties in ML models lead to a drop in predictive performance. This is due to penalizing the sensitive variables to control its contributions towards prediction. However Fair-TabNet has overall improved on both predictive and fairness performance compared to TabNet. The proposed loss terms in Fair-TabNet  introduce a regularization effect in the model as well as a disentanglement inductive bias that could explain the improvement in predictive performance compared with TabNet. Table \ref{tab:sensitive_performance} gives a summary on the fairness performance of the compared models. The  definition of the fairness metrics are given in Table \ref{tab:fair-definition}. For all the metrics being close to zero is the best in terms of fairness except for DIR where the metric is close to 100. Figure \ref{fig:sens_repr} depicts a T-SNE plot of Fair-TabNet sensitive representation where the colors indicates different sub-categories of the sensitive variables. The learned representation ${\bf r}_s$ is able to cluster well the sensitive variables. The loss function $\mathcal L_{sens}$ was trained with a sensitive target variable with the following sub-categories: Female-Foreigner, Female-Local, Male-Foreigner, Male-Local. These categories are used as multi-class target variable in $\mathcal L_{sens}$. We note that the sensitive representation was able to cluster the two sub-categories in gender together due to the presence of the sensitive variables in the input features. Figure \ref{fig:pred_repr} depicts a T-SNE plot of the predictive representation. The orthogonality constraint added by $\mathcal L_{diff}$  is reflected in this plot. All sensitive sub-categories are intermingled. It is not possible to identify clusters of sensitive variables. Table \ref{tab:sensitive_performance} summarises the results comparing the different models based on the fairness metrics. We also applied a t-test to assess the statistical significance of the performance difference across models obtained from the different training splits as depicted in Figures \ref{fig:metric_prediction}, \ref{fig:metric_gender} and \ref{fig:metric_nationality}. 



\begin{table}
{
\resizebox{1\linewidth}{!}{%
\begin{tabular}{lcccccc}
\toprule
& Metric & LightGBM & CatBoost & XGBoost & TabNet & Fair-TabNet\\
\midrule
& AOE &       10.16$\pm$1.1 &        1.22$\pm$1.2 &       0.75$\pm$0.3 &       0.8$\pm$0.4 &       1.35$\pm$1.2 \\
& DG-FPR &       12$\pm$0.9 &        1.02$\pm$1.4 &       0.43$\pm$0.1 &       0.57$\pm$0.5 &       0.71$\pm$0.6 \\
& DIR  &       67.7$\pm$1.9 &       72.74$\pm$7.1 &      74.34$\pm$5.4 &     74.7$\pm$10.8 &     79.92$\pm$16.5 \\
& EOD &        8.3$\pm$1.3 &        1.4$\pm$1.2 &       1.1$\pm$0.7 &       1.1$\pm$0.8 &       1.99$\pm$2.2 \\
& SPD &        13.0$\pm$0.9 &        1.5$\pm$1.7 &       0.9$\pm$0.2 &        0.9$\pm$0.6 &        1.2$\pm$0.8 \\
\bottomrule
\end{tabular}
}
}
\caption{Fairness performance of the compared models. The sensitive variable is Nationality (local vs. foreigner).}
\label{tab:sensitive_performance}
\end{table}


\section{Conclusion}
We proposed Fair-TabNet a deep learning model suited for tabular data  as an extension of TabNet model to incorporate fairness properties for the prediction appointment no-show from tabular data. Fair-TabNet learns a disentangled representation which lead to gain in predictive and fairness performance. The predictive performance of TabNet is close to the state-of-art approach and could be improved by a better tuning of hyper-parameters. 

\FloatBarrier
\bibliography{jmlr-sample}

\begin{thebibliography}{}

\bibitem[AlMuhaideb et~al., 2019]{almuhaideb2019prediction}
AlMuhaideb, S., Alswailem, O., Alsubaie, N., Ferwana, I., and Alnajem, A.
  (2019).
\newblock Prediction of hospital no-show appointments through artificial
  intelligence algorithms.
\newblock {\em Annals of Saudi Medicine}, 39(6):373--381.

\bibitem[Arik and Pfister, 2019]{arik2019tabnet}
Arik, S.~O. and Pfister, T. (2019).
\newblock Tabnet: Attentive interpretable tabular learning.
\newblock {\em arXiv preprint arXiv:1908.07442}.

\bibitem[Bousmalis et~al., 2016]{bousmalis2016domain}
Bousmalis, K., Trigeorgis, G., Silberman, N., Krishnan, D., and Erhan, D.
  (2016).
\newblock Domain separation networks.
\newblock In {\em Advances in neural information processing systems}, pages
  343--351.

\bibitem[Carreras-Garc{\'\i}a et~al., 2020]{carreras2020patient}
Carreras-Garc{\'\i}a, D., Delgado-G{\'o}mez, D., Llorente-Fern{\'a}ndez, F.,
  and Arribas-Gil, A. (2020).
\newblock Patient no-show prediction: A systematic literature review.
\newblock {\em Entropy}, 22(6):675.

\bibitem[Creager et~al., 2019]{creager2019flexibly}
Creager, E., Madras, D., Jacobsen, J.-H., Weis, M.~A., Swersky, K., Pitassi,
  T., and Zemel, R. (2019).
\newblock Flexibly fair representation learning by disentanglement.
\newblock {\em arXiv preprint arXiv:1906.02589}.

\bibitem[Gier, 2017]{gier2017missed}
Gier, J. (2017).
\newblock Missed appointments cost the us healthcare system \$150 b each year.
\newblock {\em Health Management Technology}, 2.

\bibitem[Kheirkhah et~al., 2015]{kheirkhah2015prevalence}
Kheirkhah, P., Feng, Q., Travis, L.~M., Tavakoli-Tabasi, S., and Sharafkhaneh,
  A. (2015).
\newblock Prevalence, predictors and economic consequences of no-shows.
\newblock {\em BMC health services research}, 16(1):1--6.

\bibitem[Li et~al., 2019]{li2019individualized}
Li, Y., Tang, S.~Y., Johnson, J., and Lubarsky, D.~A. (2019).
\newblock Individualized no-show predictions: Effect on clinic overbooking and
  appointment reminders.
\newblock {\em Production and Operations Management}, 28(8):2068--2086.

\bibitem[Locatello et~al., 2019]{locatello2019fairness}
Locatello, F., Abbati, G., Rainforth, T., Bauer, S., Sch{\"o}lkopf, B., and
  Bachem, O. (2019).
\newblock On the fairness of disentangled representations.
\newblock In {\em Advances in Neural Information Processing Systems}, pages
  14611--14624.

\bibitem[Louizos et~al., 2015]{louizos2015variational}
Louizos, C., Swersky, K., Li, Y., Welling, M., and Zemel, R. (2015).
\newblock The variational fair autoencoder.
\newblock {\em arXiv preprint arXiv:1511.00830}.

\bibitem[Madras et~al., 2018]{madras2018learning}
Madras, D., Creager, E., Pitassi, T., and Zemel, R. (2018).
\newblock Learning adversarially fair and transferable representations.
\newblock {\em arXiv preprint arXiv:1802.06309}.

\bibitem[Sarhan et~al., 2020]{sarhan2020fairness}
Sarhan, M.~H., Navab, N., Eslami, A., and Albarqouni, S. (2020).
\newblock Fairness by learning orthogonal disentangled representations.
\newblock {\em arXiv preprint arXiv:2003.05707}.

\bibitem[Zhang et~al., 2018]{zhang2018mitigating}
Zhang, B.~H., Lemoine, B., and Mitchell, M. (2018).
\newblock Mitigating unwanted biases with adversarial learning.
\newblock In {\em Proceedings of the 2018 AAAI/ACM Conference on AI, Ethics,
  and Society}, pages 335--340.

\end{thebibliography}
\appendix
\section[]{}
\clearpage
\onecolumn
\counterwithin{figure}{section}
\counterwithin{table}{section}

\begin{figure}
\centering
\includegraphics[width=\textwidth]{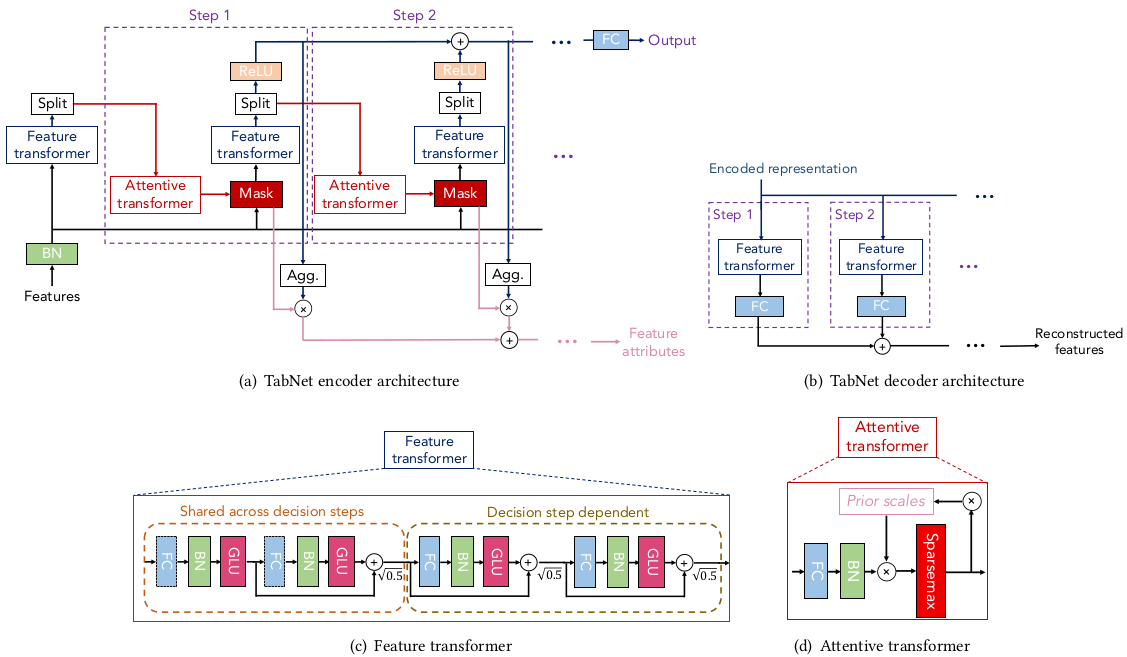}
\caption{TabNet architecture. Figure taken from \cite{arik2019tabnet}}
\label{fig:tabnet}
\end{figure}

\begin{table}
\begin{tabular}{|c|p{30mm} |l|c|}
\hline
 Metric& \textbf{Name} &\textbf{Formula}&\textbf{Best value}\\
  \hline

AOE&\textbf{Average    odds error}&  $\frac{1}{2}(|FPR_{S=1}-FPR_{S=0}|+|TPR_{S=1}-TPR_{S=0}|)$& 0\\

DG-FPR&\textbf{Difference generalized FPR}& $\mathbb{E}_{X:S=1}(\hat{Y}|Y=0)-\mathbb{E}_{X:S=0}(\hat{Y}|Y=0)$& 0\\

DIR&\textbf{Disparate Impact Ratio}&  $\frac{TPR_{S=1}}{TPR_{S=0}}$&  100\\

EOD&\textbf{Equal opportunity difference} &  $TPR_{S=1}-TPR_{S=0}$&     0\\

SPD&\textbf{Statistical parity difference}&   $Pr(Y=1|S=1)-Pr(Y=1|S=0)$&   0\\
\hline

\end{tabular}  
\caption{List of fairness metrics. $\hat{Y} $ and $Y$ are the predicted and the true classes respectively. The sensitive variable $S=1$ and $S=0$ for the deprived and privileged classes respectively. Note that generalized FPR=FPR for binary classification}
\label{tab:fair-definition}
\end{table}

\counterwithin{figure}{section}
\counterwithin{table}{section}

\begin{table}
\centering
{
\begin{tabular}{lcccccc}
\toprule
& Metric &    lightgbm &     catboost &      xgboost &     TabNet &      Fair-TabNet \\
\midrule
& AOE &   5.34$\pm$0.82 &     2.4$\pm$1.36 &     1.83$\pm$0.5 &   1.94$\pm$0.97 &    1.88$\pm$1.34 \\
& DG-FPR &   4.62$\pm$0.67 &     0.6$\pm$0.97 &     0.19$\pm$0.1 &    0.4$\pm$0.34 &    0.46$\pm$0.46 \\
& DIR &   84.3$\pm$1.74 &    70.53$\pm$4.5 &    73.23$\pm$4.3 &  70.94$\pm$7.01 &  79.64$\pm$18.88 \\
& EOD &   6.06$\pm$1.12 &     4.2$\pm$1.86 &    3.47$\pm$0.99 &   3.49$\pm$1.73 &      3.3$\pm$2.3 \\
& SPD &   5.55$\pm$0.69 &    1.43$\pm$1.17 &    0.93$\pm$0.17 &   1.12$\pm$0.53 &    0.99$\pm$0.83 \\
\bottomrule
\end{tabular}
}
\caption{Fairness metrics of the different compared models. The sensitive variable is gender.}
\end{table}

\begin{figure}
\centering
\includegraphics[width=0.8\textwidth]{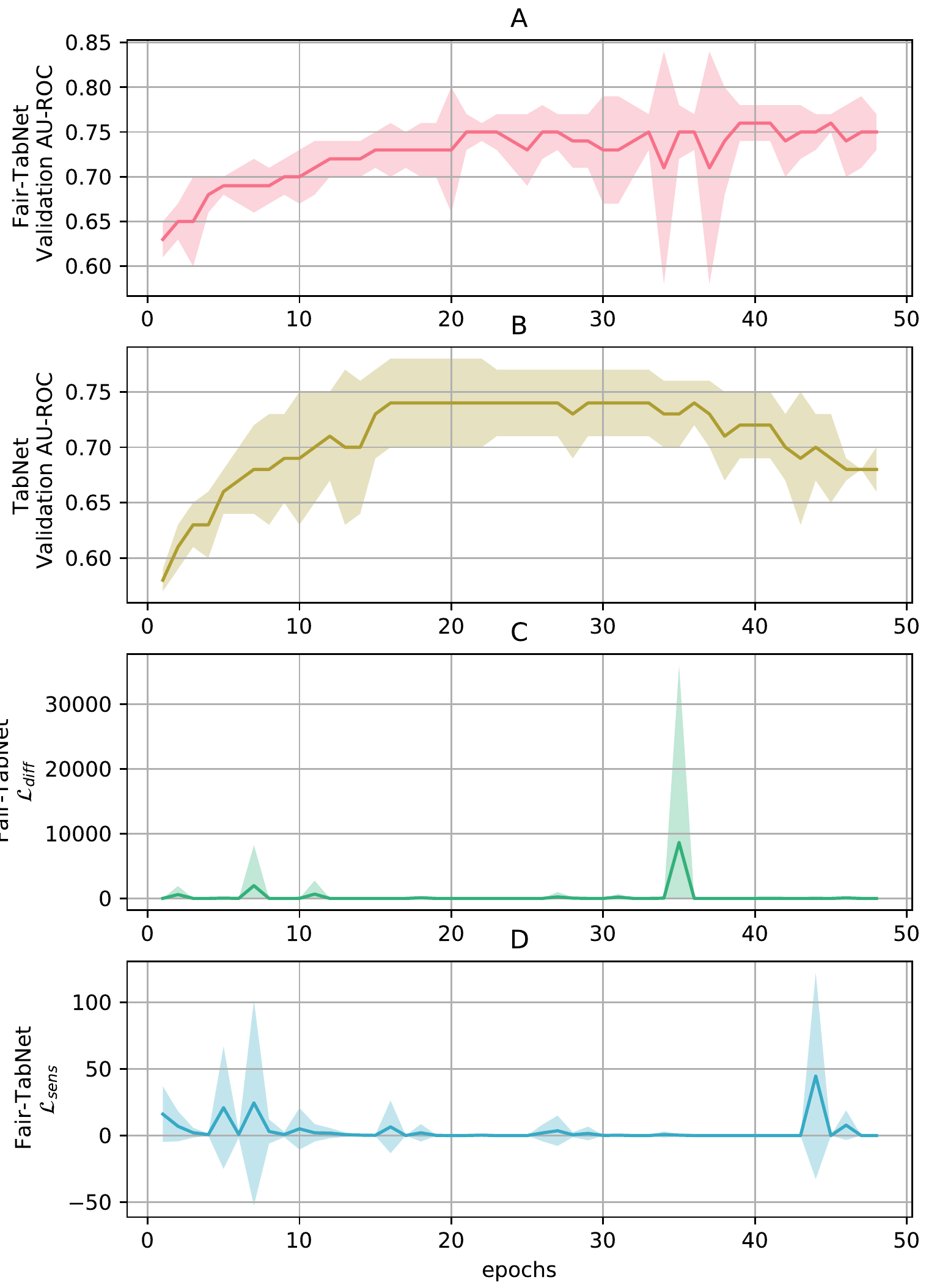}
\caption{Loss functions during Fair-TabNet training.}
\label{fig:loss}
\end{figure}

\begin{figure}
\centering
\includegraphics[width=0.7\textwidth]{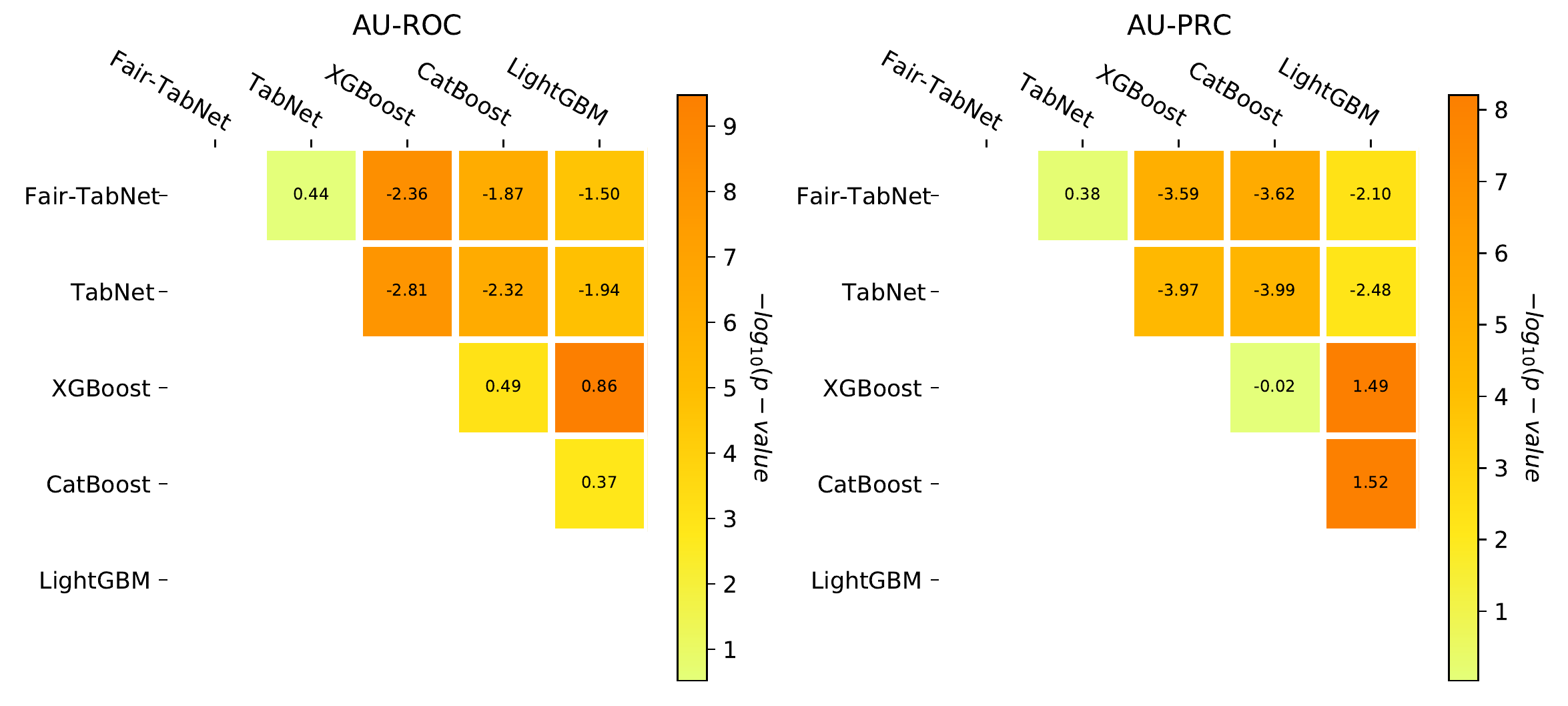}
\caption{Prediction performance comparison in terms of AU-ROC (Area Under ROC) and AU-PRC (Area Under Precision Recall). The values indicate the difference of in performance average between rows and columns. Positive values indicate that the model in row has a higher performance than model in column. The color indicates $-log_{10}(p_{value})$ of the t-test between performance on model in row and model in column.}
\label{fig:metric_prediction}
\end{figure}
 
\begin{figure}
\centering
\includegraphics[width=\textwidth]{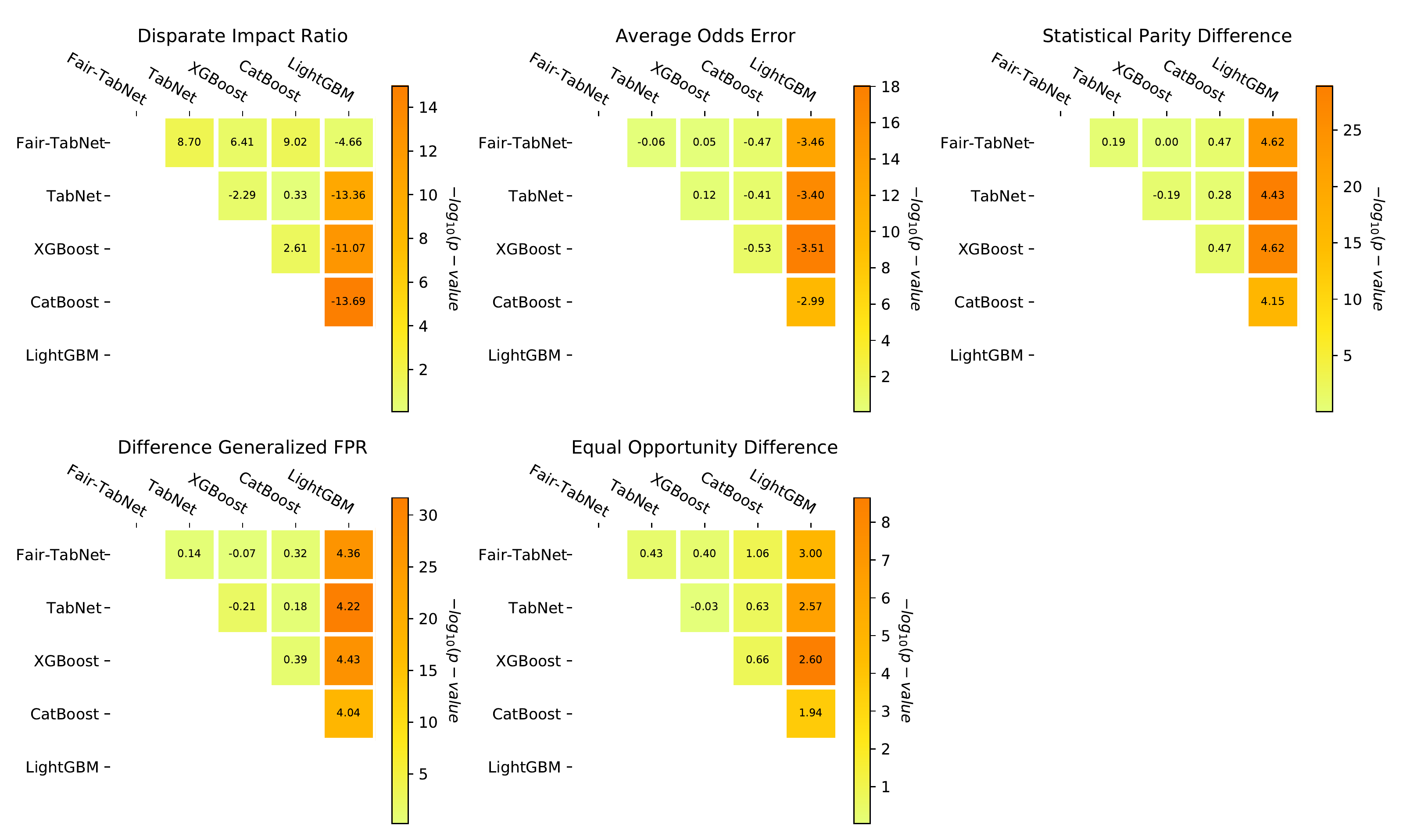}
\caption{t-test results on performance difference for the compared models. The color indicates the significance level and value show the difference of average performance of the models in rows minus the model in columns. Positive values indicate that the model in row has a higher performance than model in column.  The sensitive variable is gender (female vs. male).}
\label{fig:metric_gender}
\end{figure}
   
\begin{figure}
\centering
\includegraphics[width=\textwidth]{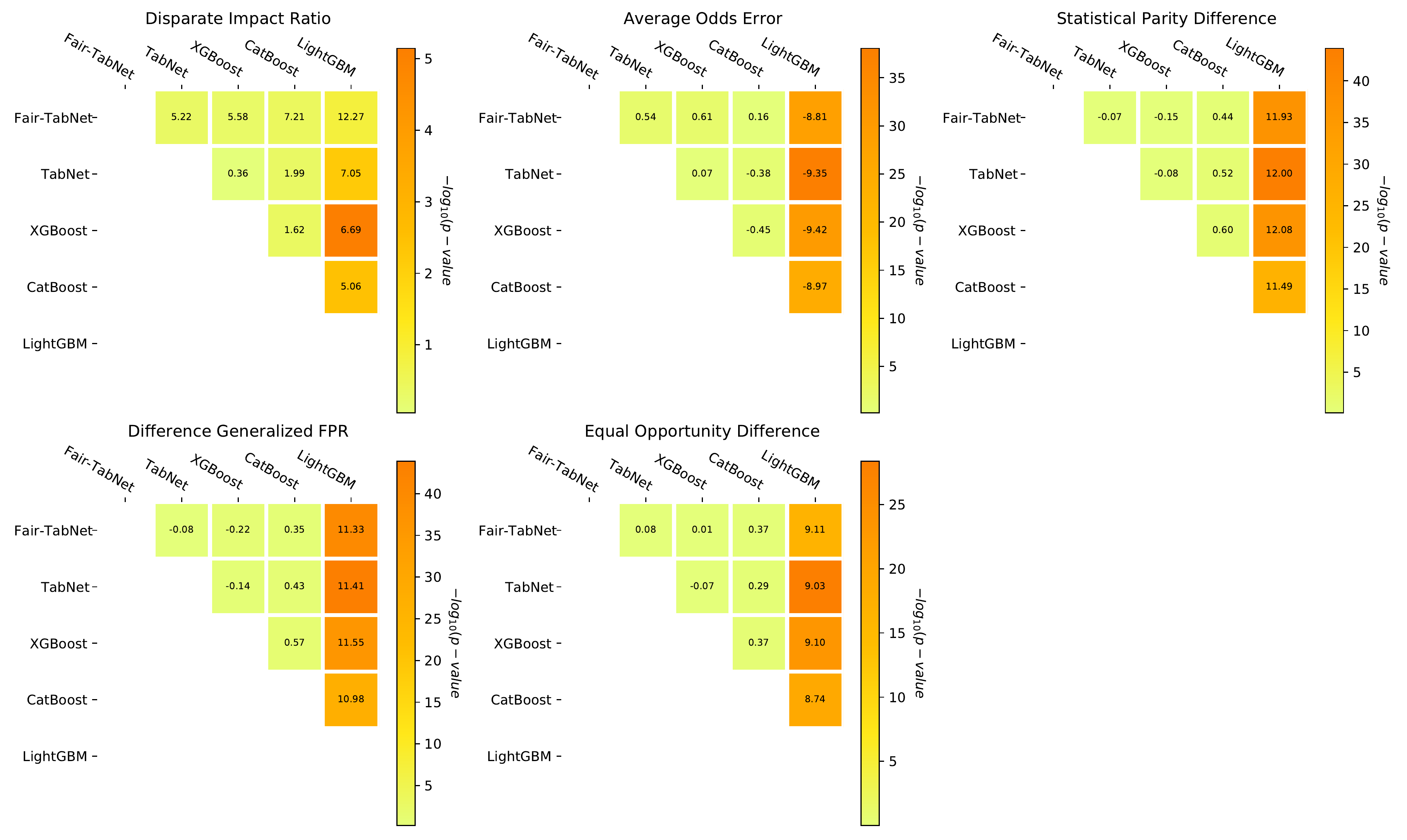}
\caption{t-test results on performance difference for the compared models. The color indicates the significance level and value show the difference of average performance of the models in rows minus the model in columns. Positive values indicate that the model in row has a higher performance than model in column. The sensitive variable is nationality (local vs. foreigner). }
\label{fig:metric_nationality}
\end{figure}
\begin{figure}
\centering
\includegraphics[width=\textwidth]{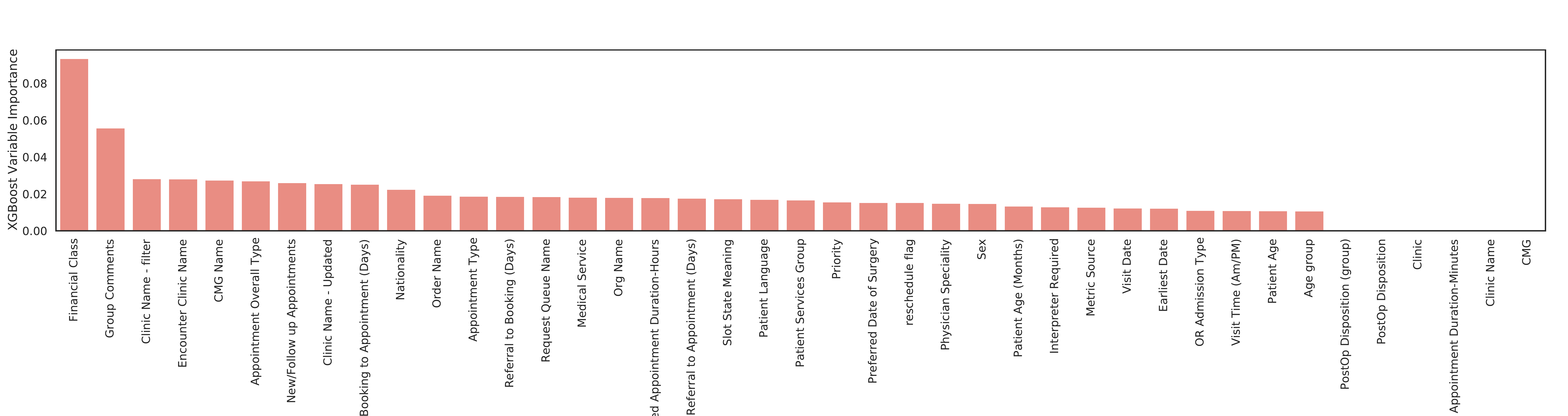}
\caption{Interpretability by Variable importance in XGBoost. The sensitive variables are highly contributing to the prediction.}
\label{fig:xgboost_interpretability}
\end{figure}

\begin{figure}
\centering
\includegraphics[width=\textwidth]{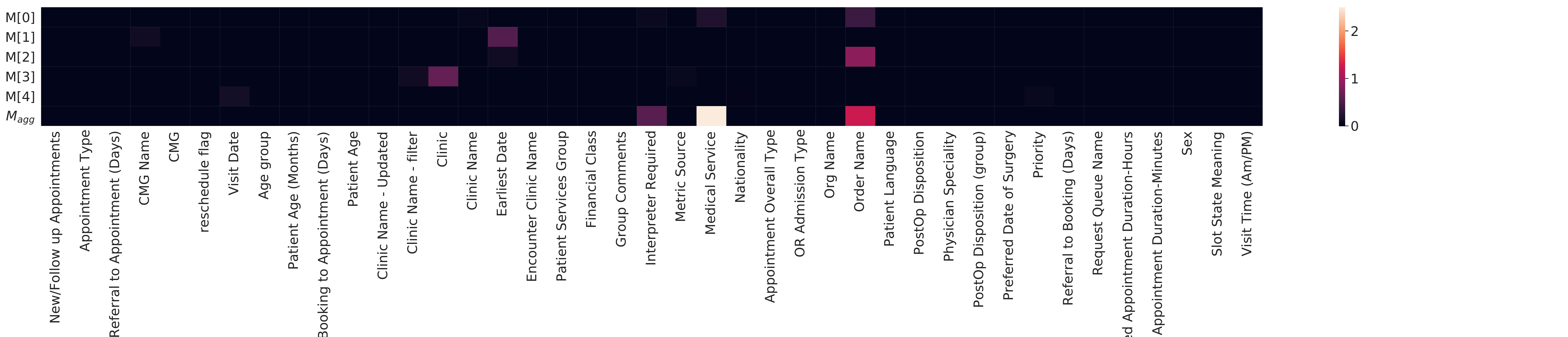}
\caption{Interpretability in Fair-TabNet from the masks and aggregated mask.}
\label{fig:fair_tabnet_interpretability}
\end{figure}


\end{document}